\documentclass[conference]{IEEEtran}
\ifCLASSINFOpdf
\else
\fi

\usepackage{cite}
\usepackage{graphicx}
\usepackage{subfig}
\usepackage{balance}
\usepackage{url}
\usepackage{xspace}
\usepackage{algorithmic}
\usepackage{algorithm}
\usepackage{multirow}

\usepackage{epsfig}
\usepackage{amssymb}
\usepackage{amsmath}
\usepackage{amsfonts}

\usepackage{placeins}

\def\ie{\emph{i.e.,~}}
\def\eg{\emph{e.g.,~}}

\newcommand{\algname}{{\sc DocSetLabeler}\xspace}
\DeclareMathOperator*{\argmax}{\arg\!\max}

\usepackage{rotating}
\usepackage{tikz}

\begin{document}
\title{Topic Similarity Networks: \\Visual Analytics for Large Document Sets}

\author{\IEEEauthorblockN{Arun S. Maiya}
\IEEEauthorblockA{Institute for Defense Analyses\\
Alexandria, VA 22311\\
Email: amaiya@ida.org}
\and
\IEEEauthorblockN{Robert M. Rolfe}
\IEEEauthorblockA{Institute for Defense Analyses\\
Alexandria, VA 22311\\
Email: rolfe@ida.org}
}


%


\maketitle

\begin{abstract}
We investigate ways in which to improve the interpretability of LDA topic models by better analyzing and visualizing their outputs.  We focus on examining what we refer to as {\em topic similarity networks}:  graphs in which nodes represent latent topics in text collections and links represent similarity among topics.  We describe efficient and effective approaches to both building and labeling such networks.  Visualizations of topic models based on these networks are shown to be a powerful means of exploring, characterizing, and summarizing large collections of unstructured text documents.  They help to ``tease out'' non-obvious connections among different sets of documents and provide insights into how topics form larger themes.  We demonstrate the efficacy and practicality of these approaches through two case studies:  1) NSF grants for basic research spanning a 14 year period and 2) the entire English portion of Wikipedia.
\end{abstract}



%
\IEEEpeerreviewmaketitle

\section{Introduction and Motivation}
\label{sec:intro}

In this paper, we study network visualizations as a means of enhancing the interpretability of probabilistic topic models for insight discovery.  We focus on what is perhaps the most popular and prevalently-used topic model:  {\em latent Dirichlet allocation} or LDA \cite{Blei2003Latent}.  Topic modeling algorithms like LDA discover latent themes (\ie topics) in document collections and represent documents as a combination of these themes.  Thus, they are critical tools for exploring text data across many domains.  Indeed, it is often the case that users must {\em discover} the subject matter buried within large and unfamiliar document sets (\eg sensemaking in text data).  Keyword searches are inadequate here, as it is unclear on where to even begin searching.  Topic discovery techniques such as LDA are a boon to users in such scenarios, as they reveal the content in an unsupervised and automated fashion.  Automated topic organization can potentially facilitate the comprehension of unfamiliar document data on even a massive scale.

However, it is often quite challenging to obtain a ``big picture'' view of the larger trends in a document collection from only the raw output of an LDA model.  LDA is fundamentally a statistical tool that returns a probability distribution for each document showing the relative presence (or absence) of various discovered topics.  These topics, in turn, are represented as probability distributions over words (typically unigrams).  Words with the highest estimated probabilities for a discovered topic are used as a {\em label} for the topic.  Exploring text corpora using only these raw outputs is considerably challenging.  In order to derive insights and identify larger trends within the document collection, one is left to inspect these numerical distributions, which can be difficult, non-trivial, and far from straightforward.  The problem is exacerbated as document collections under consideration grow.  For instance, with the existence of scalable, MapReduce implementations of LDA (\eg \cite{Zhai2011Using,Wang2009PLDA}), it is now possible to train an LDA model on massive text corpora with many latent topics (\ie big data).  The inferred topics discovered by these LDA implementations, can themselves pose their own unique data challenge.  It is often unclear on how best to effectively browse these topics to discover information of interest.  This, in fact, tends to be a significant challenge even for large data (as opposed to ``big data'') --- \eg document collections on the order of tens of thousands or hundreds of thousands.  In the present work, we investigate the use of what we refer to as {\em topic similarity networks} to address these challenges.  {\em Topic similarity networks} are graphs in which nodes represent latent topics in text collections and links represent similarity among topics.  We describe efficient and effective methods to both building and labeling such networks.   

~\\
{\bf Summary of Contributions.}  Our contributions in both the areas of topic visualization and topic labeling are summarized below.

\begin{enumerate}
\item \emph{Constructing Topic Similarity Networks:}  In Section \ref{sec:topicsim}, we describe the construction of {\em topic similarity networks}, our approach to big data visualization.  We exploit these networks to discover how topics form larger themes.  We employ the use of community detection in network visualizations to discover such macro-level themes including the sometimes subtle connections among these themes. 

\item \emph{Labeling Topic Similarity Networks:}  In Section \ref{sec:labeling}, we describe an approach to expressively labeling discovered topics.  Our method, based on keyphrase extraction, is purely unsupervised, extractive, and demonstrably efficient.  These labels are, then, employed as node labels in our {\em topic similarity networks} to enable better characterization of large document sets.  It is surprising to note that, to the best of our knowledge, few of the existing works on topic visualization (discussed in the next section) make use of automated topic labeling methods. Our work, then, represents one of the first examinations of the efficacy of automated topic labeling in actual topic visualizations of large, real-world data.
\end{enumerate}

There has been a wave of recent work to address challenges in both visualizing topics and labeling topics -- each of which we discuss separately in light of our work.

\section{Background and Related Work}
\label{sec:background}

\subsection{Visualizing Topics}
\label{sec:intro.vis}

A number of both graphical and text-based visualizations and user interfaces have been proposed in the existing literature to browse topics (\eg  \cite{Crossno2011TopicView,Gretarsson2012TopicNets,Chuang2012Interpretation,
Chaney2012Visualizing,Maiya2013Exploratory,Eisenstein2012TopicViz,Wei2010TIARA}).  Several, like {\em TopicViz} by Eisenstein et al. \cite{Eisenstein2012TopicViz} and {\em TopicNets} by Gretarsson et al.\cite{Gretarsson2012TopicNets}, are quite innovative and make significant strides towards improving the interpretability of learned topic models.  However, most of these existing methods focus on shedding light on the relationships between topics and documents (or attributes of documents).  Although some (\eg  \cite{Gretarsson2012TopicNets}) support the inference of pair-wise similarity between topics, they do not provide insights into how topics come together to form larger themes or the subtle connections among seemingly disparate groups of topics.  Such insights are important in obtaining a ``big picture'' view of ill-understood document collections.  An exception to this rule is work on {\em correlated topic models} (or CTMs) and its variants (\eg \cite{Li2006Pachinko,Chen2013Scalable,Blei2005Correlated}).  CTMs model and infer associations among topics.  These associations can be further mined to produce clusters of topics that represent larger themes for incorporation into visualizations.  These models, however, reveal certain challenges when applied in real-world scenarios.  First, existing visualizations based on CTM and its variants do not appear to easily lend themselves to extracting the kinds of insights mentioned above.  This is due both to the way in which the topic relations are constructed and depicted and also the way in which the topic nodes are labeled (topic labeling is discussed in the next section). One may refer to \cite{Li2006Pachinko,Chen2013Scalable,Blei2007Correlated} for examples of these existing visualizations and for comparison to our visualizations shown later.  Second, some approaches, such as \cite{Li2006Pachinko}, artificially constrict the topic relation structure with specification of what are referred to as supertopics, which can hinder a view of the subtle connections among different and seemingly disparate groups of topics and subtopics.  A third issue is related to practical scalability.  Chen et al. \cite{Chen2013Scalable} showed that CTM is unable to process a corpus of 285K documents in any reasonable time frame (\ie it will not finish within a week).  Similarly, an approach to infer topic hierarchies proposed by \cite{Wang2013Phrase} is limited to short texts only.  ScaCTM, a parallelized extension to CTM, was shown to be substantially more scalable given a cluster of 40 machines \cite{Chen2013Scalable}.  But, for certain domains, such machine clusters may not be available at sites of deployment. In fact, it is often the case that only a single multi-core machine is available to process millions of documents, as the storage capacity of today's machines often outstrips their processing capacity. Even in scenarios where one has access to a large machine cluster, LDA is {\em significantly} more scalable and efficient because it does not learn the correlation structure among topics.   (See \cite{Chen2013Scalable} for a time complexity analysis of CTM, ScaCTM, and LDA).   Given these aforementioned issues and the clear scalability, efficiency, and also prevalence of LDA, our objective in this work is to infer these topic associations in an organic fashion from the raw output of the {\em original} LDA model. As we will describe in Section \ref{sec:topicsim}, we do so by constructing {\em topic similarity networks}:  networks depicting the similarity (represented as links) among topics (represented as nodes).  Next, we discuss existing work on the labeling of topics.

\subsection{Labeling Topics}
\label{sec:intro.lab}

A topic similarity network is only useful as a visualization tool if the identity of network nodes are easily discernible. Several visualization schemes label topics by simply using the most probable word (or words) from the topic model (\eg \cite{Gretarsson2012TopicNets,Chen2013Scalable,Li2006Pachinko}).  However, LDA-derived labels have been observed to not always be adequately expressive of the topic (\eg see \cite{Mei2007Automatic,Wei2010TIARA,Wang2013Phrase}).  As a result, a number of methods have been proposed to better label topics in an automated fashion (\eg  \cite{Lau2011Automatic,Wang2013Phrase,Maiya2013Exploratory,
Blei2009Visualizing,Wang2007Topical,Mei2007Automatic}).  Unfortunately, for a variety of reasons, most of these existing techniques are unable to handle the large text corpora we consider in this work.  In Section \ref{sec:labeling}, we describe our own method to label topics to address gaps in this existing literature on topic labeling.  To better motivate the use of our own labeling method, we describe several goals that must be met by any labeling scheme for a topic similarity network in light of existing work on topic labeling.

~\\
\noindent
{\bf Unsupervised.}  The labeling method must be unsupervised, as obtaining a training set for a supervised labeling method can be prohibitively expensive and time-consuming.

~\\
\noindent
{\bf Extractive.}  The labeling method must be extractive.   That is, labels must be generated directly from the terms within the corpus under consideration, as opposed to an external reference corpus such as Wikipedia.  This is especially important for the government and corporate domains, which often deal with document collections describing sensitive or proprietary information, state-of-the-art ``bleeding edge'' technology, or otherwise esoteric subject matter.   Such information may not reside in publicly available reference corpora like Wikipedia.  This requirement prevents us from utilizing methods such as \cite{Lau2011Automatic}, which employs the use of reference corpora when labeling topics.

~\\
\noindent
{\bf Supportive of User Interactivity.}  Topic similarity networks are intended for use with {\em interactive} systems utilizing full-text search and faceted navigation of documents (\eg Solr search engine\footnote{\url{https://lucene.apache.org/solr}}).  Under these scenarios, the documents comprising topics may be filtered in various ways {\em after} creation of the topic model. For instance, in the government domain, only those documents containing certain markings might be deemed of interest and selected in a visualization.   Labels heavily associated with documents that have been filtered out may no longer adequately describe the remaining documents pertaining to important sub-topics.  Labeling methods that are tightly coupled with the topic model (\eg \cite{Wang2013Phrase,
Blei2009Visualizing,Wang2007Topical,Mei2007Automatic}) cannot cope well with such dynamic scenarios. Moreover, it is prohibitively expensive to re-generate the topic model on the filtered document collection.   For these reasons, our labeling method, described in Section \ref{sec:labeling}, is purposefully de-coupled from the output of LDA.  Hence, it can re-label topics in a filtered document collection without having to re-generate the topic model.  Our labeling method, then, can best be characterized as a cluster labeling approach to topic labeling.   

~\\
\noindent
{\bf Efficient.}  Dynamic filtering of document collections, as described above, also necessitates a need for efficiency in the labeling approach.  As the document collection is filtered in various ways, the labeling method might be repeatedly executed on a large document collection, which can be problematic for some existing labeling methods.  For instance, we were unsuccessful in executing the approach by \cite{Blei2009Visualizing} on the document sets of interest in this work. The approaches by \cite{Wang2013Phrase} and \cite{Mei2007Automatic} also do not appear to scale as easily or as well to larger collections of longer documents.  The method from \cite{Wang2013Phrase}, for example, was designed only for very short texts (\eg titles only).
~\\
~\\
~\\
These aforementioned issues motivate our development of a custom labeling method for use with topic similarity networks  --- a method that can scale to even massive collections of documents.  We begin a discussion of our work with a brief overview of LDA and the notation and symbols used throughout this paper.

\section{Preliminaries}
\label{sec:prelim}
Let $D =\{d_1, d_2, \ldots,d_N\}$ represent a document collection of interest and let $K$ be the number of topics or themes in $D$.   Each document is composed of a sequence of words:  $d_i=\langle w_{i1}, w_{i2}, \ldots, w_{iN_i}\rangle$, where $N_i$ is the number of words in $d_i$ and $i \in \{1 \ldots N\}$.  Let $W=\bigcup_{i=1}^{N}f(d_i)$ be the vocabulary of $D$, where $f(\cdot)$ takes a sequence of elements and returns a set.  Probabilistic topic models like LDA take $D$ and $K$ as input and produce two matrices as output.  The matrix $\theta \in \mathbb{R}^{N \times K}$ is the document-topic distribution matrix and shows the distribution of topics within each document.  The matrix $\beta \in  \mathbb{R}^{K \times |W|}$ is the topic-word distribution matrix and shows the distribution of words in each topic.  Each row of these matrices represents a probability distribution.  For any topic $i \in \{1, \ldots, K\}$, the $L$ terms with the highest probability in distribution $\beta_i$ are typically used as thematic labels for the topic.  We use these LDA-derived labels as a baseline for comparison in our work.  But first, we describe construction of the topic similarity network.

\section{Constructing the Network}
\label{sec:topicsim}

LDA captures the degree to which both documents and words are topically related.  However, relations among the topics themselves are {\em not} explicitly captured. As we will show shortly, such topic-level relations can be used to construct network representations of text corpora.  These representations, in turn, can be used to better understand, characterize, and visualize the themes in a document collection.  In the present work, we define these relations based on topic similarity.

~\\
\noindent
{\bf Measuring Topic Similarity.}  Recall that topics are represented as probability distributions over vocabulary $W$ and captured by the matrix $\beta$.  Thus, the similarity for any two topics can be directly computed by comparing the word distributions from $\beta$.  The Kullback-Leibler (KL) divergence, a distance measure of two probability distributions, is often used to make such comparisons (\eg \cite{Gretarsson2012TopicNets}, \cite{Wei2010TIARA}).  However, KL divergence satisfies neither the triangle inequality nor symmetry and is, therefore, not a metric.  As such, it is less appropriate for defining network links based on similarity (the complement of distance).   Although symmetric versions of KL divergence exist, we instead employ the Hellinger distance metric to compute topic similarity.  Specifically, for any two topics $x,y \in \{1 \ldots K\}$, the Hellinger similarity is measured as:
\begin{equation}\label{eq:hell}
H_S(\beta_x,\beta_y) = 1-\frac{1}{\sqrt{2}}\sqrt{\sum^{|W|}_{i=1}(\sqrt{\beta_{xi}}-\sqrt{\beta_{yi}})^2 }.
\end{equation}

A topic similarity network $G=(V,E)$ can be constructed where $V=\{v_1 \ldots v_K\}$ is the set of nodes representing discovered topics and $E$ is the set of edges representing similarities among topics.  For any two topics $x,y \in \{1 \ldots K\}$, an edge $\{v_x,v_y\} \in V$ exists if and only if $H_S(\beta_x,\beta_y)$ is greater than some pre-defined threshold, $\xi$.  

~\\
\noindent
{\bf Measuring Topic Similarity in MapReduce.}  Note that, when constructing a topic similarity network as just described, the number of computed similarities scales quadratically with $K$.  However, since $K \ll |D|$, the method remains computationally viable even for very large document collections. Moreover, with some well-placed substitutions, $\beta$ can be represented using a sparse matrix format for efficient in-memory processing of massive document sets. (We currently employ a compressed sparse row format for storing and manipulating $\beta$.)   Nevertheless, for scenarios when even sparse representations of $\beta$ are unwieldy and a high degree of parallelization is desired, we propose a MapReduce implementation of the topic similarity computation.  When breaking down problems into distributable units of work under the MapReduce model for parallelization, key-value pairs are employed as the core data structure \cite{Dean2008MapReduce}.  In our case, each cell in the matrix $\beta$ can be represented as a key-value pair of the form ($i$ : ($j$, $\beta_{ji}$)), where $i \in \{1 \ldots |W|\}$ is the index of a word (\ie column), $j \in \{1 \ldots K\}$ is the index of the topic (\ie row), and $\beta_{ji}$ is the probability of word $i$ appearing in topic $j$.  If grouping by key, we obtain a key-value representation of each column in $\beta$.  That is, the values list for
 any key $i \in \{1 \ldots |W|\}$ comprises the set of tuples $\{(j, \beta_{ji}) \mid j \in \{1 \ldots K\}\}$). The {\em map} operation accepts these key-value pairs as input and outputs key-value pairs of the form ($x,y$ : $e_i$), where the new key $x,y \in \{1 \ldots K\}$ are pairs of topics appearing in the aforementioned values list and the value $e_i = (\sqrt{\beta_{xi}}-\sqrt{\beta_{yi}})^2$, for each word $i \in \{1 \ldots |W|\}$.  Thus, the {\em map} operation completes the inner expression for Hellinger similarity (shown in Equation \ref{eq:hell}) for every word represented in $\beta$.  The {\em reduce} operation simply sums these values for every pair of topics and completes the Hellinger similarity computation by taking the square root of this sum, multiplying by $\frac{1}{\sqrt{2}}$, and subtracting from one.  The resultant network, constructed as described above, can be exploited to discover insights, trends, and patterns among the topics in $D$.  For the present work, we employ the use of a community detection algorithm to discover insights into how topics are related to each other and form larger themes.

~\\
\noindent
{\bf Discovering Larger Themes.}  A {\em community} can be loosely defined as a set of nodes more densely connected among themselves than to other nodes in the network \cite{Blondel2008Fast}.  Within the context of a topic similarity network, such communities should represent groups of highly-related topics, which we refer to as {\em topic groups}.  To detect these communities (or topic groups), we employ the use of the Louvain algorithm, a heuristic method based on modularity optimization \cite{Blondel2008Fast}.  Modularity measures the fraction of links falling within communities as compared to the expected fraction if links were distributed evenly in the network \cite{Newman2006Modularity}.  The algorithm initially assigns each topic node to its own community.  At each iteration, in a local and greedy fashion, topic nodes are re-assigned to communities with which it achieves the highest modularity.  As a greedy optimization method, the Louvain algorithm is exceptionally efficient and fast, even with a large number of topics.   As the authors of \cite{Blondel2008Fast} note, the computational complexity of the method is unknown, but it experimentally appears to run in $O(n\log n)$ time.  When the nodes in these constructed topic similarity networks are marked by their inferred community affiliation and labeled to express the topics they represent, the networks become powerful tools for exploration and discovery in large and heterogeneous text corpora.  We discuss labeling of topic nodes next.

\section{Labeling the Network}
\label{sec:labeling}

\begin{table*}[thb]
\centering
{\scriptsize
\begin{tabular}{l|l|l} \hline \hline
{\bf Actual Topic}         &  {\bf Labels from LDA}   & {\bf Labels from \algname}            \\ \hline
{\bf Fluid Mechanics and Fluid Dynamics}    & flow,fluid,flows,fluids,dynamics,transports   & fluid dynamics, fluid mechanics, multiphase flow                   \\ 

{\bf Game Theory}      &  agents,theory,game,agent,games,equilibrium & game theory, economic agents, repeated games   \\ 
{\bf Graph Theory}      &  discrete,graph,combinatorial,theory,combinatorics,graphs & graph theory, algebraic combinatorics, ramsey theory    \\ 
  {\bf Human Evolution}           &  modern,fossil,early,years,human,age & modern humans, human evolution, hominid evolution    \\
{\bf Hydrology}    & water,river,hydrologic,watershed,balance,surface   & hydrologic controls, watershed scale, alpine basins                   \\ 
   {\bf Modal Analysis in Structural Engineering}      &  mode,modes,research,vibration,direction,coupling & normal modes, vibration control, modal analysis    \\  
   {\bf Object Recognition}      &  object, objects,features,recognition, oriented,feature & object recognition, curved objects, cluttered scenes    \\ 
      {\bf Protein Function/Mechanisms}      &  protein,proteins,function,role,biochemical,phosphorylation & protein kinases, protein phosphorylation, protein import    \\ 
      {\bf Protein Structure}      &  protein,proteins,binding,structure,amino,acid & protein structure, protein folding, amino acid    \\ 
      {\bf Social Psychology}      &  social,people,research,individuals,attitudes,status & social psychology, social influence, social perception    \\ 
  \hline \hline
\end{tabular}
\caption{{\footnotesize {\bf [NSF Grants.]}  Ten discovered NSF topics and the highest-ranked labels assigned to each by both LDA and \algname.}}
\label{tab:nsflabels}
}
\vskip -0.1in
\end{table*}

An algorithm capable of generating expressive thematic labels for any subset of documents in a corpus can greatly facilitate both characterization and navigation of document collections.  Here, we employ such an algorithm to label nodes in a topic similarity network, as each node is a topic comprising a subset of documents in the corpus.   Our approach, referred to as \algname, is a purely unsupervised, extractive method and shown in Algorithm \ref{alg1}.\footnote{Lines $4$--$11$ of Algorithm \ref{alg1} are a variation of the KERA algorithm described in \cite{Maiya2013Exploratory}.}  \algname takes $D_S$, a subset of corpus $D$, as input, where $D_S$ consists of all documents associated with some LDA-discovered topic $t \in \{1 \ldots K\}$.  This subset can be constructed in one of two ways.  The first is to populate $D_S$ with all documents $d_i$ (where $i \in \{1 \ldots N\}$) for which the topic proportion $\theta_{it}$ is greater than some pre-defined threshold (\eg $0.3$ was used in \cite{Wei2010TIARA}).  The second is to construct $D_S$ by transforming topics into mutually-exclusive clusters, where the topic cluster for document $d_i$ is $\argmax_{x}{\theta_{ix}}$.  We employ the latter approach, as it better eliminates noise contributed by foreign topics (\ie $\{1 \ldots K\} - \{t$\}).  Labels for topic $t$ are, then, extracted by \algname directly from the text constituting the documents in $D_S$.

\begin{algorithm}[htb]
\caption{\algname algorithm}
\label{alg1}
{\footnotesize
\begin{algorithmic}[1]
\REQUIRE $D_S\subset D$, a subset of corpus $D$
\REQUIRE $C$, the number of candidate terms to consider
\REQUIRE $L$, the number of labels to return for document set ($L\leq C$)
\REQUIRE $\mathrm{stopwords}$, list of terms to filter out
\STATE $\mathrm{pos}$ =  a hash table
\STATE $\mathrm{neg}$ =  a hash table
\FORALL{$d \in D$}
\STATE $\mathit{terms1}=\mathrm{extractSignificantPhrases}(d,\mathrm{stopwords} )$
\STATE $\mathit{terms2}=\mathrm{extractNounPhrases}(d, \mathrm{stopwords})$
\STATE $\mathit{terms3}=\mathrm{extractProperNounUnigrams}(d, \mathrm{stopwords})$
\STATE $\mathit{candidates} =(\mathit{terms1}\cap \mathit{terms2})\cup \mathit{terms3}$
\FORALL{$c \in \mathit{candidates}$}
\STATE $x = \text{normalized frequency of term } $c$ \text{ in } d$
\STATE $y = 1- \frac{\text{index of first occurrence of } c \text{ in }d}{\text{num. of words in d}}$
\STATE (weight of term $c$) $=\frac{2\cdot x\cdot y}{x+y}$
\ENDFOR
\IF{ $d \in D_S$}
\STATE $\mathrm{pos[d]}$ = top $C$ terms based on weight
\ELSE
\STATE $\mathrm{neg[d]}$ = top $C$ terms based on weight
\ENDIF
\ENDFOR
\FORALL{$\ell \in \bigcup_{x\in\mathrm{pos.values()}}x$}
\STATE \# compute information gain for each label $\ell$
\STATE (score of label $\ell$) $= \mathrm{calcScore}(\ell, \mathrm{pos}, \mathrm{neg}  )$
\ENDFOR
\STATE $\mathit{top\_candidates} =$ top $C$ labels based on information gain
\STATE \# optionally re-sort final top candidates
\STATE $\mathit{top\_candidates} = \mathrm{re\_sort}(\mathit{top\_candidates})$
\STATE return top $L$ labels from $\mathit{top\_candidates}$
\end{algorithmic}
}
\end{algorithm}

\algname is essentially a {\em descriptive} model of topic labeling that follows naturally from four observed characteristics of high-quality, topic-representative labels:  {\em Expressivity}, {\em Prominence}, {\em Prevalence}, and {\em Discriminability}.

~\\
\noindent
{\bf Expressivity.}  {\em Expressivity} captures the extent to which labels express and represent themes.  Previous works have noted that human-assigned labels tend towards multi-word noun phrases, as they are more expressive than unigrams (\eg see \cite{Turney2000Learning}).  
The term ``information retrieval,'' for instance, is more expressive than just ``information'' or ``retrieval'' alone.  Unigrams tend to most often be expressive when denoting uniqueness (\ie a proper noun). This is especially true of research reports, our domain of interest, as proper noun unigrams denote important concepts, systems, techniques, or programs (\eg ``LinearSVM,'' ``F-22'').  Lines 4-6 in Algorithm \ref{alg1} explicitly extract terms conforming to the above principles.  Noun phrases\footnote{We use the POS pattern: {\sc (adjective)*(noun)+}.} and proper nouns are extracted using {\em hunpos}, an open-source, HMM-based, part-of-speech tagger.\footnote{\url{http://code.google.com/p/hunpos/}}  The $\mathrm{extractSignificantPhrases(\cdot)}$ function uses likelihood ratio tests to extract phrases of multiple  words that occur together more often than chance.\footnote{This is known as {\em collocation extraction} \cite{Manning1999Foundations}.}  For a bigram of words  $w_1$ and $w_2$, this association, $\mathit{assoc}(\cdot,\cdot)$, is measured as: 

\begin{equation}
\mathit{assoc}(w_1,w_2) = 2\sum_{ij}n_{ij}\log\frac{n_{ij}}{m_{ij}},
\end{equation}

where $n_{ij}$ are the observed frequencies of the bigram from the contingency table for $w_1$ and $w_2$ and $m_{ij}$ are the expected frequencies assuming that the bigram is independent \cite{Dunning1993Accurate}. Only phrases with a p-value less than $0.001$ are extracted.  These tests can also be used to measure associations of words within n-grams where $n\geq 3$ (\eg trigrams).  However, we limit phrases to the $n<3$ cases to save space in the visualizations. 

~\\
~\\
\noindent
{\bf Prominence.}  {\em Prominence} captures the degree to which labels are featured prominently within individual documents.  Intuitively, prominent terms tend to make their first appearance earlier and also appear more frequently. Thus, we weight candidate labels by both frequency and position using the harmonic mean, as shown in Line 11 of Algorithm \ref{alg1}.

~\\
\noindent
{\bf Prevalence and Discriminability.}  Good labels for a particular topic appear in many documents pertaining to that topic ({\em Prevalence}) and appear rarely in other un-related topics ({\em Discriminability}).  This was also recently observed by \cite{Danilevsky2013KERT} and \cite{Wei2010TIARA}.  The concept of {\em information gain} from the field of information theory simultaneously captures both prevalence and discriminability.  Consider a document collection $D$ where documents belong to either a positive or negative category.  The {\em entropy} $\mathrm{H}$ of $D$ measures impurity as follows:
$\mathrm{H}(D)= -p^{+}\log_2(p^{+})-p^{-}\log_2(p^{-})$, where $p^{+}$ and $p^{-}$ are the proportions of positive and negative documents in $D$,      respectively.\footnote{Note that $\log_2(0)$ is taken to be $0$.}   For instance, if all documents are positive (or negative), $\mathrm{H}(D)=0$,      while a perfectly even split of positive and negative documents has entropy of $1$.  In Algorithm \ref{alg1}, we assign $D_S$ as positive and $\overline{D_S}$ as negative. The information gain $\mathrm{IG}$ of a candidate label $\ell$ in $D$, then, is the expected entropy reduction due to segmenting on $\ell$:  $\mathrm{IG}(\ell, D) = \mathrm{H}(D) - (\frac{|D^{\ell}|}{|D|}\mathrm{H}(D^{\ell}) + \frac{|\overline{D^{\ell}}|}{|D|}\mathrm{H}(\overline{D^{\ell}}) ),$  where $D^{\ell}$ is the set of documents in $D$ from which label $\ell$ was extracted.  Thus, labels with the highest information gain for $D_S$ are expected to be simultaneously common in $D_S$ (prevalence) and rare in $\overline{D_S}$ (discriminability).  Information gain is computed by the $\mathrm{calcScore(\cdot)}$ function in Algorithm \ref{alg1}.  

~\\
\noindent
{\bf Final Sorting.}  At the end of the previous step, we are left with a small number of candidate labels (\eg $C=5$) for each topic.  There are several options for choosing the final label for the topic node.  For instance, one could simply select the label with the highest information gain (\ie the existing sorting).  One might also select the label most frequently extracted from the documents pertaining to the topic.  Yet another option is to include word probabilities from $\beta$ into the final weighting.  All three approaches generally yield good (albeit slightly different) results.  For the present work, based on some preliminary testing, we choose to sort labels based on a combination of the latter two approaches, as indicated in Line 25 of Algorithm \ref{alg1}.  More specifically, we sort labels based on the mean of the normalized frequency and the combined $\beta$ probabilities for each word comprising the label.

~\\
To conclude, we briefly comment on the efficiency and scalability of our current \algname implementation. Note that, in Algorithm \ref{alg1}, Lines $1$--$12$ process documents in an online fashion and can be easily parallelized. Computing information gain also scales well to larger collections of longer documents, as it is a simple computation of different combinations of independent and dependent variables.  Moreover, it deals with a substantially reduced representation of the data (\ie generally, $C \ll N_i$ for all $i\in \{1 \ldots N\}$).  For these reasons, it is fairly straightforward to implement \algname in a variety of different parallel processing models (\eg MapReduce, multi-core processing).  Lines $1$--$12$, for instance, can be implemented as a map-only job with either zero reducers or an identity reducer.  On the other hand, for execution on single-node, multi-core, shared-memory systems (as opposed to clusters), documents can be processed in an online fashion and passed to as many processors available on the system. 

\section{Case Study 1:  NSF Research Grants}
\label{sec:nsf}

\begin{figure*}[htb]
\begin{center}
\centerline{\fbox{\includegraphics[scale=0.79, angle=0]{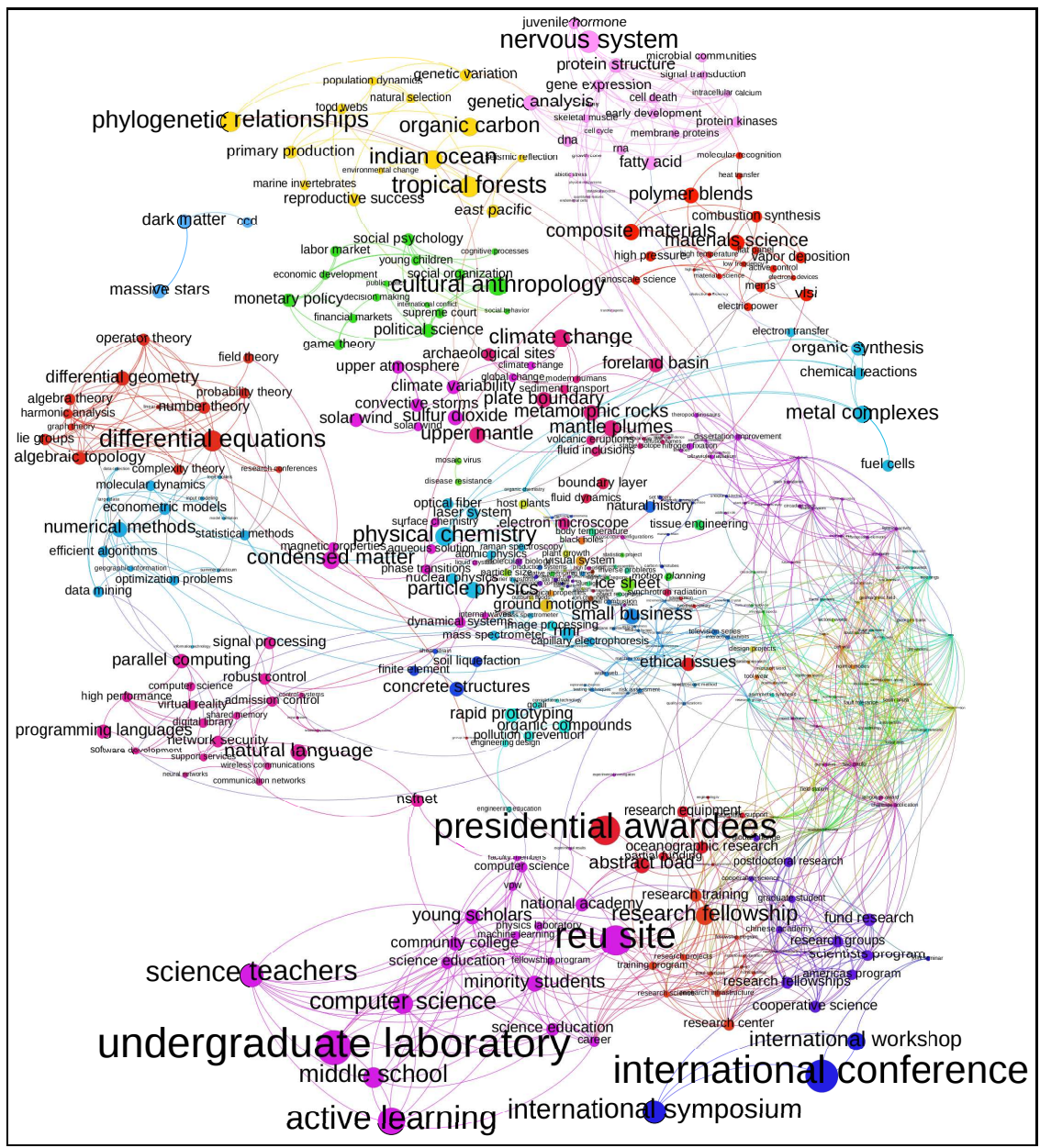}}}
\caption{{\footnotesize {\bf [NSF Grants.]} The {\em Topic Similarity Network} of 14 years of NSF research and support (\ie a total of $132,372$ research grants).  Major research topics are shown including their subtle connections to each other. Also displayed (towards the bottom of network) are major funding efforts for education support and conference support.  Node sizes indicate the number of grant abstracts pertaining to the topic.  Node colors indicate the community (or topic group) affiliation, which illustrate how research topics form larger themes.   }}
\label{fig:nsf}
\end{center}
\end{figure*}

\begin{table*}[thb]
\centering
{\scriptsize
\begin{tabular}{l|l|l} \hline \hline
{\bf Actual Topic}         &  {\bf Labels from LDA}   & {\bf Labels from \algname}            \\ \hline
 {\bf BBC}      &  bbc,british,series,television,london,uk & bbc, british television, bbc radio, british actor    \\ 
 {\bf Boxing}      &  fight,title,boxing,champion,round,boxer & professional boxer, professional career, amateur boxer    \\ 
 {\bf Computers}           &  system,computer,systems,control,computers,electronic & computer science, operating system, control system    \\
 {\bf Electronic Dance Music}      &  music,dj,label,dance,artists,records & electronic music, record label, dance music   \\ 
 {\bf Probability Theory}    & data,analysis,method,methods,distribution   & probability distribution, random variables, random variable                   \\ 
   {\bf Manufacturing}      &  company,production,factory,manufacturing,plant,industry & manufacturing company, motor company, manufacturing plant    \\  
   {\bf Motorcycles}      &  motorcycle,racing,cc,race,davidson,bike & speedway rider, cc race, british motorcycle   \\ 
{\bf Summer Olympics}    & olympics,summer,medal,won,olympic,world   & summer olympics, gold medal, bronze medal                   \\ 
{\bf Tropics}      &  species,family,tropical,habitat,natural,subtropical & tropical moist, habitat loss, natural habitats    \\ 
{\bf Winter Olympics}      &  winter,world,event,olympics,won,competed & winter olympics, world championships, ski championship  \\ 
  \hline \hline
\end{tabular}
\caption{{\footnotesize {\bf [Wikipedia.]}  Ten discovered Wikipedia topics and the highest-ranked labels assigned to each by both LDA and \algname.}}
\label{tab:wikilabels}
}
\vskip -0.1in
\end{table*}

As a realistic and informative case study, we utilize our methods to characterize and visualize basic research funded by the National Science Foundation (NSF).  The corpus considered in this case study consists of 132,372 titles and abstracts describing NSF awards for basic research between the years 1990 and 2003~\cite{Bache2013UCI}.  We executed the MALLET implementation of LDA \cite{McCallum2002MALLET}  on this corpus using $K=400$ as the number of topics and $200$ as the number of iterations.  All other parameters were left as defaults.  For topic similarity, we experimentally set $\xi$ as $0.15$ to yield a graph density of approximately $0.01$.  For the labeling of topic nodes  in the network using \algname, we set $C=5$ and $L=1$. We did not find the choice of $C$ to affect results significantly.  This is possibly due to the fact that, as described previously, we prune out candidates with no statistical significance, as measured by a likelihood ratio test.

~\\
\noindent
{\bf Topic Labeling of NSF Grants.}  Table \ref{tab:nsflabels} shows the labels generated for a sample of ten discovered topics by both \algname and LDA.  As can be seen, labels produced by \algname are more expressive and representative of the true themes of each topic.  We assigned two judges to evaluate labels for all topics.  For a fair comparison, we showed six unigram labels from LDA but only three labels (mostly bigrams) from \algname for each topic.  As shown in Table \ref{tab:nsfkappa}, both judged the labels by \algname to be generally superior ($\chi^2$=$145.73$, $P$\textless$0.0001$) with an inter-judge agreement of $0.62$, as measured by Cohen's kappa coefficient.  

\begin{table}[thb]
\centering
{\scriptsize
\begin{tabular}{l|l|l} \hline
&  \algname & LDA \\ \hline
\algname & 313 & 6 \\ \hline
LDA &  23 & 29 \\ \hline
\end{tabular}
\caption{{\scriptsize  Evaluation of labels for each method on NSF grants.  Overall, both judges chose labels from \algname to be most on-point. (Poor quality topics thrown out.)}}
\label{tab:nsfkappa}
}
\vskip -0.25in
\end{table}

\begin{figure}[htb]
\begin{center}
\centerline{\fbox{\includegraphics[scale=0.15]{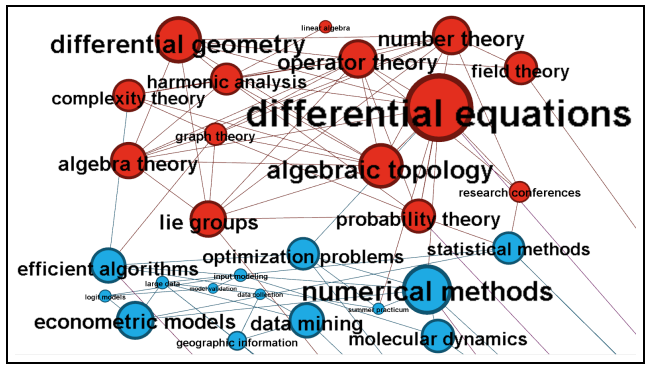}}}
\caption{{\footnotesize {\bf [NSF Grants.]}  Two discovered topic groups (or communities) pertaining to math-oriented research. The red covers pure math, while the blue is more applied.  Each are separate communities but tightly-coupled, as shown.  Together, they represent over $10,000$ documents covering a range of math subfields.}}
\label{fig:nsf-math}
\end{center}
\vskip -0.4in
\end{figure}

\begin{figure}[htb]
\begin{center}
\centerline{\fbox{\includegraphics[scale=0.13]{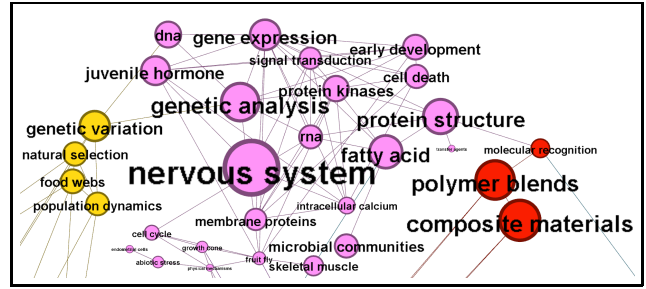}}}
\caption{{\footnotesize {\bf [NSF Grants.]}  A discovered topic group related to biology (shown in pink). Also shown are topic nodes from other related communities (\eg {\em polymer blends}, {\em population dynamics}) and their peripheral connections to this biology-related topic group.}}
\label{fig:nsf-bio}
\end{center}
\vskip -0.2in
\end{figure}

\begin{figure}[htb]
\begin{center}
\centerline{\fbox{\includegraphics[scale=0.15]{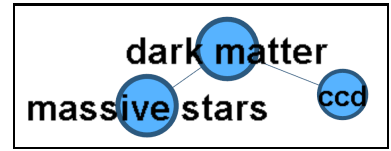}}}
\caption{{\footnotesize {\bf [NSF Grants.]}  A connected component of astronomical research topics separated from the larger network.}}
\label{fig:nsf-astro}
\end{center}
\vskip -0.4in
\end{figure}

~\\
~\\
\noindent
{\bf Visualizing NSF Grants.} A topic similarity network was constructed, with each node representing a topic and labeled using the highest ranked term returned by \algname. The network concisely presents a comprehensive and holistic view of 14 years of NSF-funded research and can be navigated and explored using any available network visualization software (\eg Gephi, Cytoscape ).  The entire network is shown in Figure \ref{fig:nsf}, where both expected and unexpected trends are revealed.  As can be seen, the visualization encapsulates the major research funding efforts for scientific research in addition to the subtle connections among them.  Major funding efforts for education and conference support are also displayed (towards the bottom).  In this network and all networks shown in this paper, node sizes indicate the number of documents pertaining to the topic represented by the node.\footnote{Although we could have sized nodes based on funding amount of the grant, we instead size nodes based on the number of documents for the sake of consistency.}  Node colors indicate the community (or topic group) affiliation.  Using this network, one can better understand how topics form larger themes, discover and characterize information of interest, and derive insights into how best to search and explore the corpus further.  It is difficult to quantitatively evaluate visualization schemes such as this.  Thus, we present illustrative examples of the patterns and trends discovered using our topic similarity network.  Figure \ref{fig:nsf-math} shows one small corner of the ``topic universe'' --- a ``social clique'' of math topics discovered by community detection within the larger network of all topics. Note that each node in the network represents hundreds of documents (or more).  Thus, this visualization of math topics clearly and concisely summarizes over $10,000$ documents.   Such visualizations also provide insights into relations between topic groups.  For instance, Figure \ref{fig:nsf-bio} shows a community of biology-related topics (shown in pink).  Here, we see peripheral connections to another life science theme (shown in yellow) containing topics such as {\em genetic variation},  {\em population dynamics}, and {\em food webs}.  We also see a peripheral connection to a material science theme (shown in red), illuminating research areas dedicated to developing materials based on biological and organic components and also the mutual interest in molecular recognition.  As a final example, Figure \ref{fig:nsf-astro} shows a connected component of astronomical research topics that appears separate from the larger network.  This last example illustrates one possible way to use these visualizations to identify outliers (\ie topics that are comparatively more different than the larger corpus based on their set of similarity scores).\footnote{While it is possible to re-connect singleton nodes to whichever node it is most similar, we have not done so in any of the presented visualizations.}

\section{Case Study 2:  Wikipedia}
\label{sec:wiki}

For our second case study, we apply our method to visualize Wikipedia topics.  The corpus considered here was obtained from the University of Alberta and comprises the entire English portion of Wikipedia.\footnote{Shaoul, C. \& Westbury C. (2010) The Westbury Lab Wikipedia
Corpus, Edmonton, AB: University of Alberta (downloaded from
\url{http://www.psych.ualberta.ca/~westburylab/downloads/westburylab.wikicorp.download.html})}  It contains over 3.3 million documents spanning a range of different topics.   We executed the MALLET implementation of LDA \cite{McCallum2002MALLET}  on this corpus using $K=1000$ as the number of topics and $200$ as the number of iterations.  All other parameters were left as defaults.  For topic similarity, we experimentally set $\xi$ as $0.2$ to yield a graph density of approximately $0.01$.  For the labeling of topic nodes  in the network using \algname, we again set $C=5$ and $L=1$.

~\\
\noindent
{\bf Labeling Wikipedia Topics.}  Table \ref{tab:wikilabels} shows a sample of ten Wikipedia topics and the labels generated for each by both LDA and \algname.   As we did with the NSF grants, we conducted a user evaluation of the labels generated for all Wikipedia topics by both LDA and our method.  From the results shown in Table \ref{tab:wikikappa}, we again see that \algname outperforms LDA ($\chi^2$=$426.68$, $P$\textless$0.0001$) with an inter-judge agreement of $0.71$, as measured by Cohen's kappa coefficient..  However, we also see that LDA performs significantly better here than on the NSF grants.  We elaborate on this observation further in Section \ref{sec:limitations}.

\begin{table}[thb]
\centering
{\scriptsize
\begin{tabular}{l|l|l} \hline
&  \algname & LDA \\ \hline
\algname & 545 & 74 \\ \hline
LDA &  30 & 199 \\ \hline
\end{tabular}
\caption{{\footnotesize  {\bf [Wikipedia.]}  Evaluation of labels for each method on Wikipedia.  Overall, both judges chose labels from \algname to be most on-point. (Poor quality topics thrown out.)}}
\label{tab:wikikappa}
}
\vskip -0.25in
\end{table}

\begin{figure}
  \centering

  \subfloat[Sports-Themed Topic Group]{\label{fig:wiki-sport}\centerline{\fbox{\includegraphics[width=0.45\textwidth]{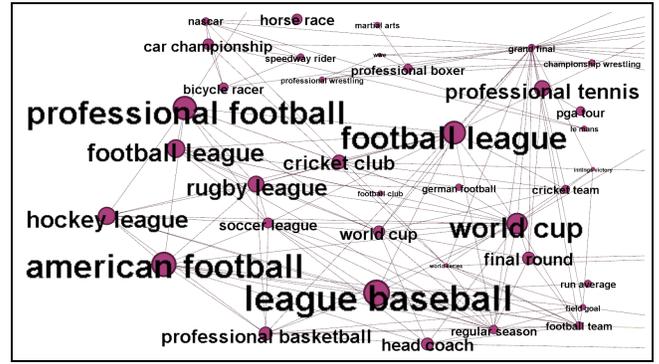}}}} \\\vspace{.05cm}
      
 \subfloat[Music/Dance-Themed Topic Group] {\label{fig:wiki-music}\centerline{\fbox{\includegraphics[width=0.45\textwidth]{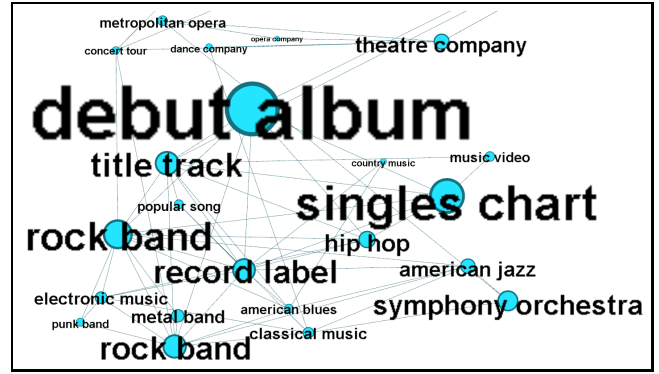}}}}
  
\caption{{\footnotesize {\bf [Wikipedia.]}  Discovered Wikipedia topic groups for:  (a) {\em Sports} and (b) {\em Music/Dance}.}}
  \label{fig:clustlab}
  \vskip -0.15in
\end{figure}

\begin{figure}[htb]
\begin{center}
\centerline{\fbox{\includegraphics[scale=0.16]{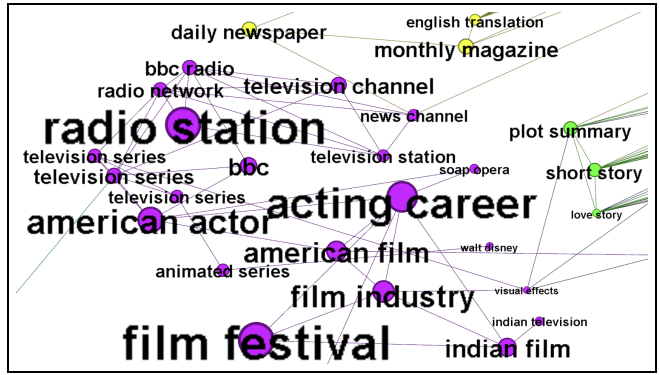}}}
\caption{{\footnotesize {\bf [Wikipedia.]}  A discovered topic group pertaining to {\em Television/Film/Radio} (shown in purple).  Also shown are the peripheral connections to topic nodes from other related communities (\eg {\em plot summary} and {\em love story} from a writing theme in green, {\em daily newspaper} and {\em monthly magazine} from a news media theme in yellow).}}
\label{fig:tvfilm}
\end{center}
\vskip -0.4in
\end{figure}

~\\
\noindent
{\bf Visualizing Wikipedia.}  A topic similarity network was constructed for Wikipedia, with nodes labeled using the highest ranked label generated from \algname for each topic.  Due to space constraints, we do not present the entire Wikipedia topic similarity network in this paper.  Rather, we provide illustrative examples of some of the major trends discovered by our method.  Two of the most salient and well-defined topic groups (\ie macro-level themes)  emerging from our visualization are {\em sports} and {\em music/dance}, shown in Figures \ref{fig:wiki-sport} and \ref{fig:wiki-music}, respectively.  We posit that this is due to the fact that authorship and editing of Wikipedia articles are crowd-sourced and the subjects of {\em sports} and {\em music/dance} both have enormous fan bases.  It should follow that television and film should also appear as salient topic groups, and this is precisely what we see in Figure \ref{fig:tvfilm}.  Also shown in Figure \ref{fig:tvfilm} are the peripheral connections to topic nodes from other related communities (\eg {\em plot summary} and {\em love story} from a writing theme in green, {\em daily newspaper} and {\em monthly magazine} from a news media theme in yellow).

\section{Limitations}
\label{sec:limitations}
In both our two case studies, \algname was observed to outperform LDA on topic labeling tasks.  However, comparing the two case studies, we see the performance differential was less for Wikipedia topics and greater for the highly technical and scientific topics present in the NSF grants corpus.  We attribute this to the fact that Wikipedia is an encyclopedia with many topics that are very general and broad in nature.  On those topics that are so broad and general that they are best summarized with a single word (\eg {\em songs}, {\em tennis}, {\em BBC}), LDA performs quite well -- albeit sometimes less well than \algname. In cases where there is not an equivalently expressive bigram (\ie two-word phrase) or proper unigram, LDA will perform better than our method, since \algname currently focuses only on bigrams and {\em proper} unigrams.  One example of the latter case is the {\em motorcycle} topic in Wikipedia shown in Table \ref{tab:wikilabels}. The top-ranked labels generated by \algname are simply not as expressive as the simple label ``motorcycle'' produced by LDA.  Addressing such cases is an area for future work.  However, we find these cases to be in the minority -- especially with respect to mining content from scientific and technical documents, which is our current and primary area of interest.

A second limitation is related to short texts.  Both LDA and \algname are optimized for articles, summaries, and reports, such as the corpora considered in this work.  Shorter documents such as abstracts are also handled well by both algorithms, as evidenced by performance on the NSF grant abstracts.  However, extremely short texts can cause difficulties.  This was observed to a certain degree in some Wikipedia topics containing many so-called ``stub'' articles of only a single sentence\footnote{It appears that Wikipedia now recommends a  minimum of three sentences for an article.  See \url{http://en.wikipedia.org/wiki/Wikipedia_talk:One_sentence_does_not_an_article_make}} (\eg one-sentence descriptions of minor fictional characters, small towns, or persons of minor notability).  One solution might be to replace LDA and \algname with algorithms specifically designed to handle short texts such as Twitter-LDA \cite{Zhao2011Comparing} and keyword extraction algorithms designed for short snippets of text \cite{Li2010Keyword}.  We leave an investigation of this for future work.

\section{Conclusion}

We have investigated the use of {\em topic similarity networks} as a practical approach to improving the interpretability of LDA topic models.   We described both how to construct such networks and an approach to labeling nodes in the network.  These methods were combined and employed to effectively characterize and explore 14 years of NSF-funded basic research and the English portion of Wikipedia using network analysis.  For future work, we plan on incorporating these visualizations into a larger, facet-based, text analytic system previously developed for the U.S. Department of Defense (see \cite{Maiya2013Exploratory} for more details on this system).

\balance

\begin{thebibliography}{10}

\bibitem{Bache2013UCI}
K.~Bache and M.~Lichman.
\newblock {UCI} machine learning repository, 2013.

\bibitem{Blei2005Correlated}
David~M. Blei and John~D. Lafferty.
\newblock {Correlated Topic Models}.
\newblock In {\em NIPS}, 2005.

\bibitem{Blei2007Correlated}
David~M. Blei and John~D. Lafferty.
\newblock {A correlated topic model of Science}.
\newblock {\em Annals of Applied Statistics}, 1(1):17--35, August 2007.

\bibitem{Blei2009Visualizing}
David~M. Blei and John~D. Lafferty.
\newblock {Visualizing Topics with Multi-Word Expressions}, July 2009.

\bibitem{Blei2003Latent}
David~M. Blei, Andrew~Y. Ng, and Michael~I. Jordan.
\newblock {Latent Dirichlet Allocation}.
\newblock {\em J. Mach. Learn. Res.}, 3(4-5):993--1022, March 2003.

\bibitem{Blondel2008Fast}
Vincent~D. Blondel, Jean-Loup Guillaume, Renaud Lambiotte, and Etienne
  Lefebvre.
\newblock {Fast unfolding of communities in large networks}.
\newblock {\em Journal of Statistical Mechanics: Theory and Experiment},
  2008(10):P10008+, July 2008.

\bibitem{Chaney2012Visualizing}
Allison Chaney and David~M. Blei.
\newblock {Visualizing Topic Models}.
\newblock In {\em ICWSM '12}, 2012.

\bibitem{Chen2013Scalable}
Jianfei Chen, June Zhu, Zi~Wang, Xun Zheng, and Bo~Zhang.
\newblock {Scalable Inference for Logistic-Normal Topic Models}.
\newblock In {\em NIPS 2013: Neural Information Processing Systems Conference},
  December 2013.

\bibitem{Chuang2012Interpretation}
Jason Chuang, Daniel Ramage, Christopher Manning, and Jeffrey Heer.
\newblock {Interpretation and Trust: Designing Model-driven Visualizations for
  Text Analysis}.
\newblock In {\em Proceedings of the SIGCHI Conference on Human Factors in
  Computing Systems}, CHI '12, pages 443--452, New York, NY, USA, 2012. ACM.

\bibitem{Crossno2011TopicView}
P.~J. Crossno, A.~T. Wilson, T.~M. Shead, and D.~M. Dunlavy.
\newblock {TopicView: Visually Comparing Topic Models of Text Collections}.
\newblock In {\em Tools with Artificial Intelligence (ICTAI), 2011 23rd IEEE
  International Conference on}, pages 936--943. IEEE, November 2011.

\bibitem{Danilevsky2013KERT}
Marina Danilevsky, Chi Wang, Nihit Desai, Jingyi Guo, and Jiawei Han.
\newblock {KERT: Automatic Extraction and Ranking of Topical Keyphrases from
  Content-Representative Document Titles}, June 2013.

\bibitem{Dean2008MapReduce}
Jeffrey Dean and Sanjay Ghemawat.
\newblock {MapReduce: Simplified Data Processing on Large Clusters}.
\newblock {\em Commun. ACM}, 51(1):107--113, January 2008.

\bibitem{Dunning1993Accurate}
Ted Dunning.
\newblock {Accurate Methods for the Statistics of Surprise and Coincidence}.
\newblock {\em Comput. Linguist.}, 19(1):61--74, March 1993.

\bibitem{Eisenstein2012TopicViz}
Jacob Eisenstein, Duen~H. Chau, Aniket Kittur, and Eric Xing.
\newblock {TopicViz: Interactive Topic Exploration in Document Collections}.
\newblock In {\em CHI '12 Extended Abstracts on Human Factors in Computing
  Systems}, CHI EA '12, pages 2177--2182, New York, NY, USA, 2012. ACM.

\bibitem{Gretarsson2012TopicNets}
Brynjar Gretarsson, John O'donovan, Svetlin Bostandjiev, Tobias~H. ̈Llerer,
  Arthur Asuncion, David Newman, and Padhraic Smyth.
\newblock {TopicNets: Visual Analysis of Large Text Corpora with Topic
  Modeling}.
\newblock {\em Journal ACM Transactions on Intelligent Systems and Technology
  (TIST)}, 3(2), February 2012.

\bibitem{Lau2011Automatic}
Jey~H. Lau, Karl Grieser, David Newman, and Timothy Baldwin.
\newblock {Automatic Labelling of Topic Models}.
\newblock In {\em Proceedings of the 49th Annual Meeting of the Association for
  Computational Linguistics: Human Language Technologies - Volume 1}, HLT '11,
  pages 1536--1545, Stroudsburg, PA, USA, 2011. Association for Computational
  Linguistics.

\bibitem{Li2006Pachinko}
Wei Li and Andrew McCallum.
\newblock {Pachinko Allocation: DAG-structured Mixture Models of Topic
  Correlations}.
\newblock In {\em Proceedings of the 23rd International Conference on Machine
  Learning}, ICML '06, pages 577--584, New York, NY, USA, 2006. ACM.

\bibitem{Li2010Keyword}
Zhenhui Li, Ding Zhou, Yun~F. Juan, and Jiawei Han.
\newblock {Keyword Extraction for Social Snippets}.
\newblock In {\em Proceedings of the 19th International Conference on World
  Wide Web}, WWW '10, pages 1143--1144, New York, NY, USA, 2010. ACM.

\bibitem{Maiya2013Exploratory}
Arun~S. Maiya, John~P. Thompson, Francisco~L. Lemos, and Robert~M. Rolfe.
\newblock {Exploratory Analysis of Highly Heterogeneous Document Collections}.
\newblock In {\em Proceedings of the 19th ACM SIGKDD International Conference
  on Knowledge Discovery and Data Mining}, KDD '13, pages 1375--1383, New York,
  NY, USA, 2013. ACM.

\bibitem{Manning1999Foundations}
Christopher~D. Manning and Hinrich Sch\"{u}tze.
\newblock {\em {Foundations of Statistical Natural Language Processing}}.
\newblock MIT Press, Cambridge, MA, USA, 1999.

\bibitem{McCallum2002MALLET}
Andrew~K. McCallum.
\newblock {MALLET: A Machine Learning for Language Toolkit}, 2002.

\bibitem{Mei2007Automatic}
Qiaozhu Mei, Xuehua Shen, and ChengXiang Zhai.
\newblock {Automatic Labeling of Multinomial Topic Models}.
\newblock In {\em Proceedings of the 13th ACM SIGKDD International Conference
  on Knowledge Discovery and Data Mining}, KDD '07, pages 490--499, New York,
  NY, USA, 2007. ACM.

\bibitem{Newman2006Modularity}
M.~E.~J. Newman.
\newblock {Modularity and community structure in networks}.
\newblock {\em Proceedings of the National Academy of Sciences},
  103(23):8577--8582, June 2006.

\bibitem{Turney2000Learning}
PeterD Turney.
\newblock {Learning Algorithms for Keyphrase Extraction}.
\newblock 2(4):303--336, 2000.

\bibitem{Wang2013Phrase}
Chi Wang, Marina Danilevsky, Nihit Desai, Yinan Zhang, Phuong Nguyen,
  Thrivikrama Taula, and Jiawei Han.
\newblock {A Phrase Mining Framework for Recursive Construction of a Topical
  Hierarchy}.
\newblock In {\em Proceedings of the 19th ACM SIGKDD International Conference
  on Knowledge Discovery and Data Mining}, KDD '13, pages 437--445, New York,
  NY, USA, 2013. ACM.

\bibitem{Wang2007Topical}
Xuerui Wang, Andrew McCallum, and Xing Wei.
\newblock {Topical N-Grams: Phrase and Topic Discovery, with an Application to
  Information Retrieval}.
\newblock In {\em Proceedings of the 2007 Seventh IEEE International Conference
  on Data Mining}, ICDM '07, pages 697--702, Washington, DC, USA, 2007. IEEE
  Computer Society.

\bibitem{Wang2009PLDA}
Yi~Wang, Hongjie Bai, Matt Stanton, Wen~Y. Chen, and Edward~Y. Chang.
\newblock {PLDA: Parallel Latent Dirichlet Allocation for Large-Scale
  Applications}.
\newblock In {\em Proceedings of the 5th International Conference on
  Algorithmic Aspects in Information and Management}, volume 5564 of {\em AAIM
  '09}, pages 301--314, Berlin, Heidelberg, 2009. Springer-Verlag.

\bibitem{Wei2010TIARA}
Furu Wei, Shixia Liu, Yangqiu Song, Shimei Pan, Michelle~X. Zhou, Weihong Qian,
  Lei Shi, Li~Tan, and Qiang Zhang.
\newblock {TIARA: a visual exploratory text analytic system}.
\newblock In {\em Proceedings of the 16th ACM SIGKDD international conference
  on Knowledge discovery and data mining}, KDD '10, pages 153--162, New York,
  NY, USA, 2010. ACM.

\bibitem{Zhai2011Using}
K.~Zhai, J.~Boyd-Graber, and N.~Asadi.
\newblock {Using Variational Inference and MapReduce to Scale Topic Modeling}.
\newblock {\em ArXiv e-prints: arXiv:1107.3765 [cs.AI]}, July 2011.

\bibitem{Zhao2011Comparing}
WayneXin Zhao, Jing Jiang, Jianshu Weng, Jing He, Ee-Peng Lim, Hongfei Yan, and
  Xiaoming Li.
\newblock {Comparing Twitter and Traditional Media Using Topic Models}.
\newblock In Paul Clough, Colum Foley, Cathal Gurrin, GarethJ Jones, Wessel
  Kraaij, Hyowon Lee, and Vanessa Mudoch, editors, {\em Advances in Information
  Retrieval}, volume 6611 of {\em Lecture Notes in Computer Science},
  chapter~34, pages 338--349. Springer Berlin Heidelberg, Berlin, Heidelberg,
  2011.

\end{thebibliography}

%
%

\end{document}